\documentclass{article}
\usepackage{ijcai24}

\usepackage{times}
\usepackage{soul}
\usepackage{url}
\usepackage[hidelinks]{hyperref}
\usepackage[utf8]{inputenc}
\usepackage[small]{caption}
\usepackage{graphicx}
\usepackage{amsmath}
\usepackage{amsthm}
\usepackage{booktabs}
\usepackage{algorithm}
\usepackage{algorithmic}
\usepackage[switch]{lineno}
\usepackage{multirow}
\usepackage{amssymb}
\usepackage{makecell}
\usepackage{color}
\usepackage{longtable}
\usepackage{booktabs}

\title{LLMs can Find Mathematical Reasoning Mistakes by Pedagogical Chain-of-Thought}

\author{
Zhuoxuan Jiang$^{1,2}$\thanks{Equal contribution and listed in alphabetical order. Zhuoxuan Jiang is the corresponding author.}
\and
Haoyuan Peng$^{2*}$\and
Shanshan Feng$^{3,4}$\and
Fan Li$^5$\And
Dongsheng Li$^6$\\
\affiliations
$^1$Shanghai Business School, Shanghai, China\\
$^2$Learnable.AI Inc., Shanghai, China\\
$^3$Centre for Frontier AI Research, A*STAR, Singapore\\
$^4$Institute of High-Performance Computing, A*STAR, Singapore\\
$^5$The Hong Kong Polytechnic University, Hong Kong, China\\
$^6$Microsoft Research Asia, Shanghai, China\\
\emails
\url{jzx@sbs.edu.cn},
\url{penghy15@gmail.com},
\url{victor_fengss@foxmail.com},
\url{fan-5.li@polyu.edu.hk},
\url{dongshengli@fudan.edu.cn}
}

\begin{document}

\maketitle

\begin{abstract}

Self-correction is emerging as a promising approach to mitigate the issue of hallucination in Large Language Models (LLMs). To facilitate effective self-correction, recent research has proposed mistake detection as its initial step. However, current literature suggests that LLMs often struggle with reliably identifying reasoning mistakes when using simplistic prompting strategies. To address this challenge, we introduce a unique prompting strategy, termed the Pedagogical Chain-of-Thought (PedCoT), which is specifically designed to guide the identification of reasoning mistakes, particularly mathematical reasoning mistakes. 
PedCoT consists of pedagogical principles for prompts (PPP) design, two-stage interaction process (TIP) and grounded PedCoT prompts, all inspired by the educational theory of the Bloom Cognitive Model (BCM). 
We evaluate our approach on two public datasets featuring math problems of varying difficulty levels. The experiments demonstrate that our zero-shot prompting strategy significantly outperforms strong baselines. The proposed method can achieve the goal of reliable mathematical mistake identification and provide a foundation for automatic math answer grading.
The results underscore the significance of educational theory, serving as domain knowledge, in guiding prompting strategy design for addressing challenging tasks with LLMs effectively.

\end{abstract}


\section{Introduction}
In recent years, Large Language Models (LLMs) have emerged as the leading force in advancing various Natural Language Processing (NLP) tasks, consistently achieving state-of-the-art performance. 
Despite the remarkable improvement in general AI abilities, LLMs still suffer from some essential problems, e.g., hallucination, which suggests that current LLMs may generate contents that look reasonable but intrinsically illogical~\cite{hallucination}. 

To overcome hallucination, many efforts have been made, among which \textit{self-correction} is one of the most recently focused directions. It empowers LLMs to correct their outputs by zero- or few-shot prompting~\cite{survey}. Some researchers treat self-correction as a single-process way~\cite{selfcorrectchen,selfcorrectmadaan}, while some researchers break down the task into two components, i.e., \textit{mistake finding} and \textit{output correction}, and achieve superior performance~\cite{bigbenchmistake}. Since finding mistakes is a fundamental and precedent skill for right correction, it is reasonable that the two-step self-correction is more reliable.



Mistake finding is a prerequisite of self-correction, while how to well achieve it is still an open challenge. Some studies show that current LLMs struggle to find mistakes reliably using basic prompting strategies~\cite{cannot-correct-yet,bigbenchmistake}.
Most previous benchmark studies mainly adopt simple and general prompting strategies, e.g., Zero-shot CoT and its variants\footnote{They are Direct trace-level prompting, Direct step-level prompting, and CoT step-level prompting. In this paper, we follow their task settings. The details of benchmark experiments with more prompting strategies will be introduced in Section 5.}. They do not fully leverage domain-related knowledge to design prompts. Domain-related knowledge is important because there are various kinds of reasoning problems, such as word sorting, tracking shuffled objects, logical deduction and multi-step arithmetic~\cite{tasks}, which require different kinds of fine-grained reasoning skills. For example, word sorting needs the skills of text and order understanding while multi-step arithmetic needs the skills of calculation and deduction.
Hence, without domain-related knowledge, the potential of LLMs may be underestimated due to inadequately designed prompts. 

In this paper, we investigate the challenge of creating efficient prompts for a specific domain to identify reasoning errors utilizing LLMs. As a starting point, we choose the mathematical domain, a basic subject in education. On the one hand, mathematical questions queries necessitate rigorous reasoning and logic, an area where current LLMs fall short. On the other hand, finding reasoning mistakes in mathematics can provide a foundation for automatic answer grading in terms of pedagogy. To this end, we conduct interdisciplinary research on LLMs prompting strategies by adopting pedagogical knowledge within this work to achieve reliable mathematical mistake finding.



\begin{figure}
\centerline{\includegraphics[width=0.45\textwidth]{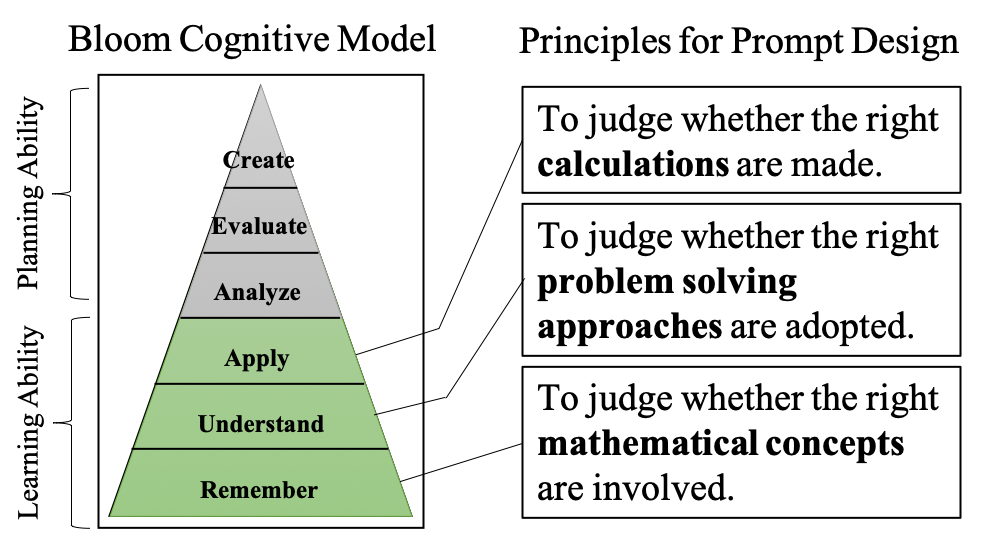}}
\caption{We develop the principles for prompt design for LLMs by leveraging the educational Bloom Cognitive Model and we focus on the \textit{learning ability}. The bold parts are keywords used in prompts.} \label{bloom}
\end{figure}


In the field of pedagogy, the teaching and learning objectives are theoretically consistent with the Bloom Cognitive Model (BCM)\footnote{Also called Bloom's Taxonomy of Educational Objectives in Cognitive Domain.}~\cite{bloom1}, which breaks the student abilities into six levels from a cognitive perspective, i.e., Remember, Understand, Apply, Analyze, Evaluate, and Create (as shown in Figure~\ref{bloom})~\cite{bloom2}. According to this theory, the first three foundational levels are associated with \textit{learning ability}, signifying that these skills are relatively easier to acquire, whereas the upper three levels involve \textit{planning ability}, which are more complex to nurture and develop.
We focus on the learning abilities in this work and leave the exploration of planning abilities as future work.

To align with BCM and bridge the gap in design prompting strategies for LLMs, we introduce a novel method named Pedagogical Chain-of-Thought (abbr. PedCoT). More specifically, the PedCoT strategy embodies two parts. First, following the BCM and the levels of \textit{learning ability}, we propose the pedagogical principles of prompt (PPP) design for LLMs. The detailed content of PPP is consistent with \textit{learning ability}, as shown in Figure~\ref{bloom}. Second, concerning the content of prompts to be grounded when interacting with LLMs, we formulate the two-stage interaction process (TIP) and grounded prompts, consistent with the three principles. 

To assess the efficacy of our approach, we gather two publicly available datasets, each containing mathematical problems of varying degrees of complexity. i.e., multi-step arithmetic and multi-step word problems. 
The experimental results, compared against strong baselines, consistently reveal a noteworthy contrast to what most existing studies have asserted. Specifically, we observe that current LLMs, such as GPT-4 Turbo, can effectively find mathematical reasoning mistakes through our proposed PedCoT. The result highlights the importance and value of domain knowledge in prompting the reasoning abilities of LLMs.


The main contributions of this paper include:

\begin{itemize}
\setlength{\itemsep}{0pt}
  \setlength{\parskip}{0pt}
  \setlength{\itemindent}{0em}
    \item We conduct an interdisciplinary study and investigate a new problem on how to leverage domain knowledge to guide prompt design for LLMs, i.e., leveraging pedagogical theories to find mathematical mistakes.
    \item We develop a novel zero-shot prompting strategy named PedCoT to bridge the gap between educational theory and prompts for LLMs, which is featured by (1) pedagogical principles for prompt (PPP) design, (2) two-stage interaction process (TIP) and (3) PedCoT prompts.
    \item Experiments on two public datasets with various complexity of math problems demonstrate that contrary to what most existing studies claim, current LLMs actually can find mathematical reasoning mistakes by using our PedCoT equipped with pedagogical theory.
\end{itemize}





\section{Related Work}

\subsection{Automatic Answer Grading}

The task of finding mistakes among mathematical reasoning steps is analogous to the task of automatic answer grading (ASG) from the perspective of pedagogy. The ASG has garnered significant attention within the educational technology community. 
The methods for ASG are manifold, which depend on specific subjects and/or question types, e.g., automatic scoring for essay~\cite{essay}, reading comprehension~\cite{reading}, open math response~\cite{open_response} and short math answering grading~\cite{short_answer}. The majority of these studies focus on improving semantic modeling by incorporating the most recent language models from the Natural Language Processing (NLP) community. As to the mathematical grading task, besides capturing the semantics, evaluating the rationale (intermediate reasoning steps) on the word problem step-by-step is still much more challenging~\cite{gsm8k,naturalprover}. The emergence of LLMs significantly improves the level of machine intelligence and provides a potential direction for finding mistakes. However, the difference between ASG and mistake finding is that the former can utilize the standard answer information while the later cannot.
To the best of our knowledge, this paper represents the first attempt to leverage the pedagogical theories from step-by-step grading of mathematical answers for LLMs.

\begin{figure*}
\centerline{\includegraphics[width=0.85\textwidth]{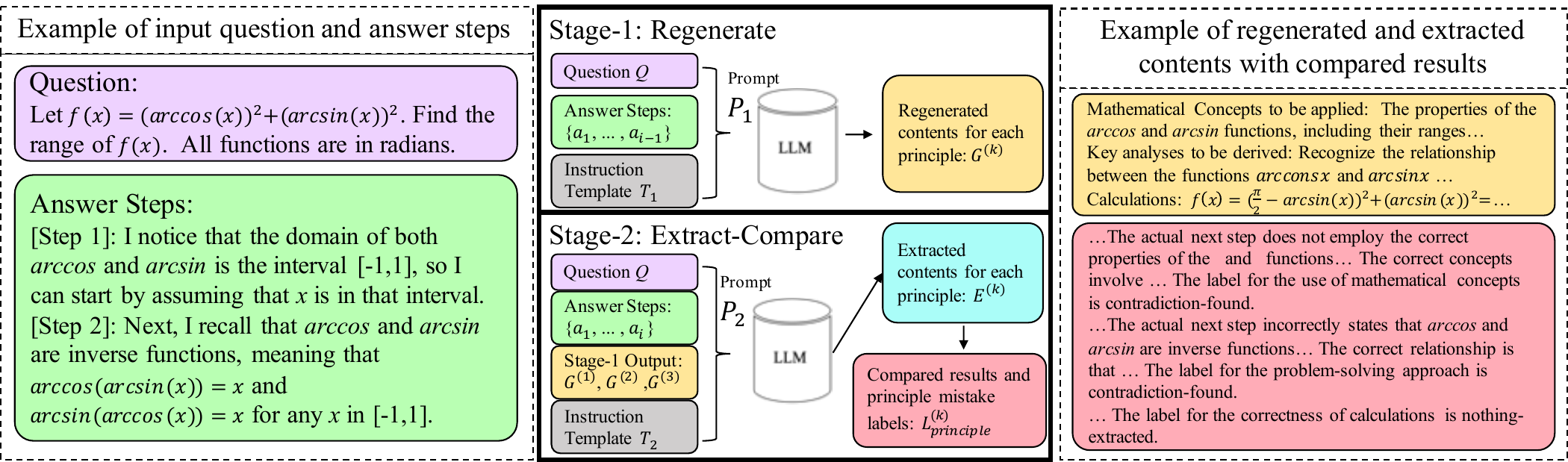}}
\caption{Diagram of two-stage interaction process (TIP) with LLMs for finding mistakes at the $i$-th step. The left and right parts are the exampling input and output contents. The detailed contents, as well as complete prompts, of the example can be referred to Appendix.} \label{twostage}
\end{figure*}

\subsection{Chain-of-Thought Prompting}

LLMs indeed shown great potential in various tasks, but prompting them efficiently can be challenging. Recently, the concept of Chain-of-Thought (CoT) has been introduced~\cite{cot}. Then various zero- and few-shot CoT prompting methods are studied to demonstrate the ability of LLMs to solve challenging tasks~\cite{self-consistency}. For example, in terms of zero-shot prompting, Zero-shot CoT simply adds `Let's think step by step' after each question, which enables LLMs to generate step-by-step rationales and achieve better final answers~\cite{0shot-cot}. Later, Plan-and-Solve Prompting instructs LLMs to devise a plan by breaking down the entire task into smaller ones and then executing each subtask to accomplish the plan, addressing the step missing and calculation issues which Zero-shot CoT may suffer from~\cite{plan-and-solve}. Furthermore, AutoCoT automatically identifies the most similar questions from a corpus and uses Zero-shot CoT to generate several demonstrations to support few-shot learning on LLMs~\cite{autocot}.

Besides zero-shot prompting, CoT methods can be enhanced by manually providing a few exemplars as part of the prompts. This approach can more effectively harness the reasoning capabilities of LLMs~\cite{react}. For example, few-shot CoT methods are proved effective by transforming the natural language-formed rationales to executable Python codes~\cite{pot,PAL}. Least-to-Most Prompting decomposes complex reasoning problems into a series of simpler sub-problems and provides specific exemplars to teach LLMs~\cite{least-to-most}. Recently, a new exemplar selection scheme has been proposed, showing that choosing good exemplars with complex reasoning CoTs is important~\cite{complexcot}. Our work is in line with the challenging zero-shot CoT prompting while we focus on finding reasoning mistakes, instead of solving math problems.




\subsection{Detecting Reasoning Errors with LLMs}

Self-correction is one of the recently focused topics among zero- and few-shot prompting methods to tackle hallucination problems~\cite {selfcorrectsaunders,selfcorrectchen,selfcorrectmadaan}. It can be divided into two steps: \textit{mistake finding} and \textit{output correction}. The former step is regarded as a fundamental skill and normally addressed by using some few-shot prompting methods~\cite{fewshotverification,fwoshotdeductive}. Nevertheless, recent studies demonstrate that the current LLMs still struggle to identify mistakes even in objective, unambiguous cases~\cite{cannot-correct-yet,bigbenchmistake}. Hence, some studies indicate that even the state-of-the-art LLMs perform poorly in fixing their own reasoning errors without external feedback and LLMs are only able to improve their outputs in terms of style and quality~\cite{rci,reflexion}. 


A recent study proposed SelfCheck~\cite{selfcheck}, which promotes LLMs to detect reasoning mistakes in a zero-shot manner. SelfCheck can partially address the problem of detecting reasoning mistakes of LLMs, while it still has some problems: (1) SelfCheck may be faced with a bias when the regenerated and real current steps to be compared are consistently wrong. To deal with the bias, PedCoT considers more comprehensive and reliable factors, including pedagogical theories, various mistake types, and different mistake labels; (2) SelfCheck requires five LLM requests to complete a check, and it is inefficient; and (3) SelfCheck is evaluated by its performance in detecting reasoning mistakes without comparing with other related methods. We conduct more comprehensive experiments on various CoT prompting methods and achieve the state-of-the-art performance.



\begin{table*}
\centerline{\includegraphics[width=0.85\textwidth]{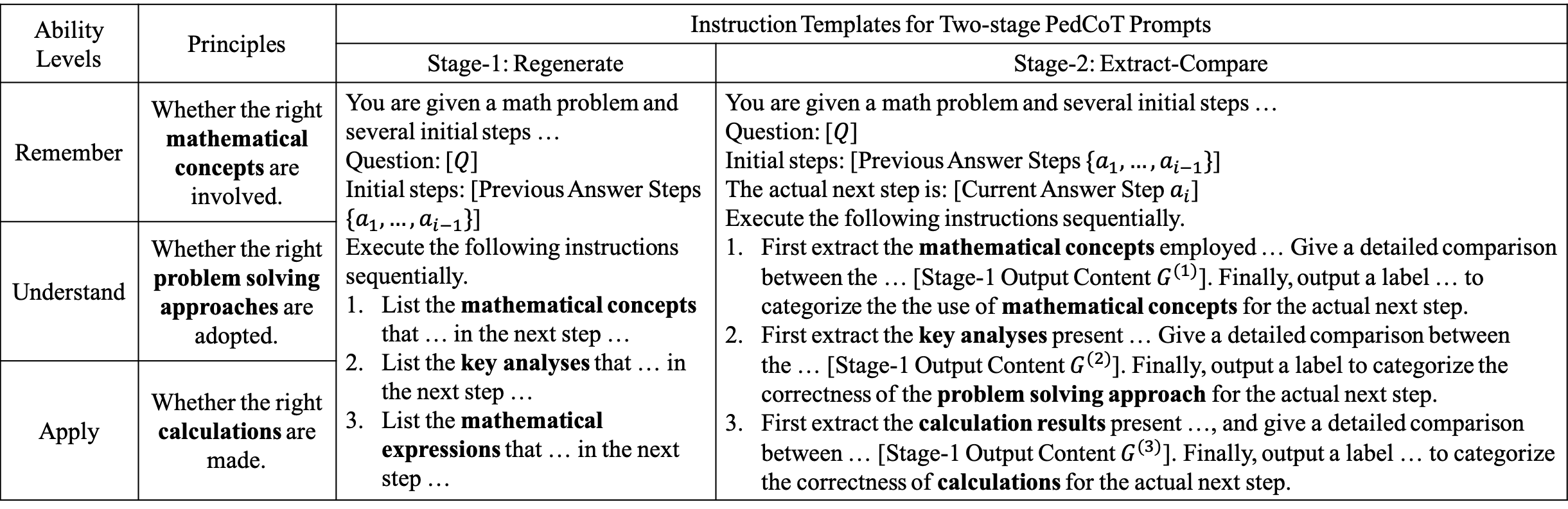}}
\caption{The proposed PedCoT prompting method which evolved from left two columns of pedagogical ability levels and principles for prompt design. We design the instruction templates for PedCoT prompts that coordinate with the two-stage interaction process (TIP). The bold parts are keywords to embody the corresponding educational meanings. We refine the wording of PedCoT prompts to avoid misunderstanding by LLMs through numerous empirical attempts. The contents in [] are variables.} \label{promts}
\end{table*}

\section{Problem Statement and Analysis}

In this section, we will firstly provide a detailed definition of the research problem. Subsequently, we will undertake a comprehensive analysis of the research problem and introduce the proposed solution to address this problem.

\subsection{Problem Definition}

We propose a novel problem of finding mathematical reasoning mistakes by zero-shot prompting LLMs. Mathematical reasoning mistakes can be divided into a three-hierarchy decision-making mistakes: namely principle mistake, step mistake, and trace mistake respectively.

Given a question (i.e., math problem) $Q$ and a trace of the top $i$ answer steps $A_i=\{a_1,...,a_i\}$, as variables, they are integrated into predefined instruction templates $T=\{T_{j}^{k}\}$ and then the combined prompts can be represented by $P=\{P_{j}\}$, where $j$ means the $j$-th stage of request on a pre-selected LLM (e.g., GPT-4) and $k$ corresponds to the $k$-th pedagogical principle. In this work, we set the $J=2$ and $K=3$ since we only need two stages of requests on LLM to complete the mistake finding task and we leverage three pedagogical principles for the time being. Therefore, at Stage-1, the input prompt $P_1$ is combined with $Q$, $A_{i-1}$ and $T_1$ to request on the LLM to obtain the intermediate content $G^{(k)}$. The process is named as Regenerate and is introduced in the following section. At Stage-2, the input prompt $P_2$ is combined with $Q$, $A_i$, $T_2$ and $G$ to request on the LLM again to extract some content $E^{(k)}$ and compare the consistency between the regenerated content and extracted content. This process is called as Extract-Compare. The LLM can finally output a principle mistake label in terms of principles which is defined as: 
$L_{principle}^{(k)}\in\{\text{correct-and-aligned, reasonable-but-incomplete, contradic-}$

\noindent$\text{tion-found, nothing-extracted}\}$,
where \textit{correct-and-aligned} means the extracted contents $E^{(k)}$ are correct and aligned with the regenerated contents $G^{(k)}$. \textit{reasonable-but-incomplete} means the extracted contents $E^{(k)}$ are partially correct but not completely aligned with the regenerated contents $G^{(k)}$. \textit{contradiction-found} means the extracted contents $E^{(k)}$ are totally contrary to the regenerated contents $G^{(k)}$. \textit{nothing-extracted} means no related content is extracted from the current answer step.

With the acquired principle mistake labels in terms of each principle, then we can obtain the 1-0 binary step mistake label for the $i$-th step defined as the following:
{\fontsize{8.0pt}{\baselineskip}\selectfont
\begin{equation}
L_{step}^{(i)} =
\begin{cases}
1, &\text{if}\quad\forall k\in\{1,2,3\}, st. L_{principle}^{(k)}\ne\text{contradiction-found}, \\
0, &\text{if}\quad\exists k\in\{1,2,3\}, st. L_{principle}^{(k)}=\text{contradiction-found}.
\end{cases} \nonumber
\end{equation} \nonumber
}

With the acquired step mistake labels in terms of each answer step, then we can obtain the 1-0 binary trace mistake label defined as the following:
{\fontsize{8.0pt}{\baselineskip}\selectfont
\begin{equation}
L_{trace} =
\begin{cases}
1, & \text{if}\quad\forall i\in\{1,...,n\}, st. L_{step}^{(i)}=1, \\
0, & \text{if}\quad\exists i\in\{1,...,n\}, st. L_{step}^{(i)}=0,
\end{cases} \nonumber
\end{equation} \nonumber
}
where $n$ is the number of steps in an answer trace. Actually, the answer steps are in a sequential order, and the step mistake labels are obtained orderly. Once the first step mistake is found, the trace is identified as mistake found and the iteration can stop. We believe that the above proposed principle mistake label schema is aligns well with educational objectives for answer grading. It provides explainable information for understanding the logic of mistake identification, as opposed to the binary judgment employed by most existing methods.




\subsection{Analysis}

The basic idea on addressing the problem of finding mathematical reasoning mistakes is to leverage pedagogical theories and bridge the gap between pedagogy and LLMs. Pedagogical theories are essential in the field of education as they provide a framework for understanding how and why students learn. They are widely used to assist teachers in evaluating the students' learning outcomes. Hence, it is believed that the pedagogical theories can be transferred to instruct LLMs in mistake finding and self-correction. Following the framework of pedagogical theories, we first develop the principles for prompt design, and then the grounded prompts can be designed in a systematical manner. In this paper, we develop three pedagogical principles for prompt (PPP) design according to ability levels that are defined by Bloom Cognitive Model. The detail analysis of the developed PPP is discussed as below:

\paragraph{Ability Level.} As shown in Figure~\ref{bloom}, six ability levels are defined by the Bloom Cognitive Model which explains the teaching objectives from the cognitive perspective of pedagogy. From bottom (low) to up (high), they are Remember, Understand, Apply, Analyze, Evaluate, and Create. Along with the theory, the lower three levels are called \textit{learning ability} which are easier to obtain, while the higher three levels, \textit{planning ability}, are notably more complex to nurture and develop. This paper focuses on the lower three ability levels.

\paragraph{Pedagogical Principles for Prompt Design.} To fill the gap between pedagogical theories and grounded prompts for LLMs to execute, we propose three pedagogical principles for prompt (PPP) design, corresponding to the three ability levels. The PPP can guide the later grounded prompt design on what aspects should be involved. The right part of Figure~\ref{bloom} shows the content of principles. Admittedly, based on the BCM, each ability level contains various aspects of fine-grained skills, and the proposed principles cannot fully reflect them. For example, remembering \textit{mathematical concepts} is just a representative skill among 14 counterparts at the level of Remember. The detail of ability levels can be referred to the theory~\cite{bloom2}. As a starting point, we propose to defer the exploration of more intricate principles to subsequent research endeavours.




\section{Methodology}






Contrary to existing methodologies that solely rely on the inherent capabilities of LLMs to identify errors - an approach that often proves ineffective - we instead take a pioneer study to utilize external domain knowledge on the targeted problem. 
In this paper, we design the novel PedCoT prompts for mathematical reasoning mistake finding based on the previously developed PPP design. The PedCoT prompts are also coordinated with the newly-proposed two-stage interaction process (TIP) on LLMs. We introduce the TIP first.

\subsection{Two-Stage Interaction Process}

Assume to find mistake at the $i$-th answer step, Figure~\ref{twostage} shows the two-stage interaction process with LLMs, including Stage-1 Regenerate and Stage-2 Extract-Compare.

\paragraph{Stage-1: Regenerate.} Given the prompt $P_1$ for Stage-1, which is a combination of a math question $Q$, the previous answer steps $A_{i-1}=\{a_1,...,a_{i-1}\}$ and the instruction template $T_1$, LLMs regenerate the potential mathematical concepts, key analysis (i.e., problem-solving approaches), and calculation results (i.e., calculations) about the expected current step, denoted as $G\in\{G^{(k)}\}$. We use regular expressions to split the output contents $G$ into three segments to obtain each $G^{(k)}$. Unlike the majority of existing approaches that necessitate LLMs to explicitly regenerate the current step and compare it with the actual current step, our method can generate a pedagogical analysis of the step while keeping the actual current step, $a_i$, unseen to LLMs for the sake of objectivity.

\paragraph{Stage-2: Extract-Compare.} Given the prompt $P_2$ for Stage-2, which is a combination of a math question $Q$, the current answer steps $A_{i}=\{a_1,...,a_{i}\}$, the outputs $G$ from Stage-1 Regenerate and the instruction template $T_2$, LLMs firstly extract the real mathematical concepts, key analysis 
(i.e., problem-solving approaches), and calculation results (i.e., calculations) within the current step $a_i$, denoted as $E\in\{E^{(k)}\}$.
Then, LLMs compare the extracted contents $E$ with regenerated contents $G$ at Stage-1 along with the respective principle. At last, LLMs make the conclusions on whether the mathematical concepts, problem-solving approaches, and/or calculations are rightly involved in the current step $a_i$, i.e., giving the principle mistake label $L_{principle}^{(k)}$. Then the step mistake label $L_{step}^{(i)}$ can be acquired.



To find mistake at any step, our method only requires two requests. To the best of our knowledge, no existing other work with less than two requests can fulfil complex reasoning tasks using LLMs currently.

\subsection{Pedagogical Chain-of-Thought Prompts}

Based on pedagogical principles mentioned in Section 3.2, we develop the Pedagogical Chain-of-Thought (PedCoT) prompts for each stage of TIP. Each prompt consists of the instruction templates (i.e., $T_1$ or $T_2$) and the variables (i.e., $Q$, $A$ and/or $G$). As shown in Table~\ref{promts}, the right two columns list the grounded contents of developed prompts for each stage. The bold parts are keywords to embody the corresponding educational meanings. We refine the wording of grounded prompts to avoid misunderstanding by LLMs through numerous empirical attempts.

By utilizing the proposed PedCoT prompts, the LLMs can perform a binary assessment of the current step's accuracy. Additionally, they can provide explanatory feedback for educational purposes by generating a textual principle mistake label for each ability level. Thus, our method can bridge the gap between pedagogical theories and prompt design for LLMs, and fulfil the task of step-by-step reasoning mistake finding in the mathematical domain.


\section{Experiment}

\subsection{Datasets}

To evaluate our proposed PedCoT, we collect two public datasets containing step-level correctness labels for mathematical problems with different difficulties. 

\paragraph{BIG-Bench Mistake}~\cite{bigbenchmistake}: The original dataset consists 2,186 reasoning traces generated by PALM-2~\cite{anil2023palm}, encompassing five reasoning tasks derived from diverse domains. With an emphasis on empowering LLMs to identify reasoning errors within the realm of mathematics, we chose samples exclusively related to this field. In total, these samples encompass 1,506 reasoning steps across 300 answer traces. The type of mathematical problem consistently involves multi-step arithmetic calculations. These are relatively straightforward to solve and necessitate basic mathematical concepts and problem-solving strategies, such as the fundamental rules of arithmetic. 
62 correct traces are involved to avoid the bias that LLMs may treat every trace as faulty.

\begin{table*}
    \centering\scriptsize
    \begin{tabular}{c|clcccccccccc}
        \toprule
        Datasets & LLMs  & CoT methods & MF. Acc. & Avg. F1 & Cls. Acc. & $P_{+}$ & $R_{+}$ & $F _{+}$ & $P_{-}$ & $R_{-}$ & $F_{-}$ \\
        \hline
        \multirow{12}{*}{\makecell[c]{BIG-Bench Mistake}} & \multirow{6}{*}{\makecell[c]{GPT-4}} & Direct-Step$^\dagger$ & 42.67 & 78.67 & - & - & - & - & - & - & - \\
                  && Vanilla & 46.67 & 75.19 & 76.00 & 40.38 & 33.87 & 36.84 & 83.47 & 86.97 & 85.19      \\
                  && Zero-shot CoT  & 46.33  & 72.82  & 73.33 & 33.93 & 30.65 & 32.20 & 82.38 & 84.45 & 83.40   \\
                  && Plan-and-Solve  & 47.00  & 73.17  & 74.33 & 34.69 & 27.42 & 30.63 & 82.07 & 86.55 & 84.25   \\
                  && SelfCheck  & 81.00  & 88.37  & 88.33 & 71.43 & \textbf{72.58} & \textbf{72.00} & \textbf{92.83} & 92.44 & 92.63   \\
                  && PedCoT (ours)  &\textbf{83.00}  & \textbf{88.81}  & \textbf{89.33} & \textbf{81.25} & 62.90 & 70.91 & 90.87 & \textbf{96.22} & \textbf{93.47}   \\
        \cline{2-12}
        & \multirow{6}{*}{\makecell[c]{GPT-4 Turbo}} & Direct-Step$^\dagger$ & 43.33 & 86.24 & - & - & - & - & - & - & - \\
                  && Vanilla & 64.67  & 85.84  & 87.33 &85.29 & 46.77 & 60.42 & 87.59 & 97.90 & 92.46      \\
                  && Zero-shot CoT  & 70.00 & 88.73  & 89.67 & 89.74 & 56.45 & 69.31 & 89.66 & 98.32 & 93.79   \\
                  && Plan-and-Solve  & 77.00 & 87.33  & 88.67 & \textbf{91.18} & 50.00 & 64.58 & 88.35 & \textbf{98.74} & 93.25   \\
                  && SelfCheck  & 80.33  & 87.84  & 87.67 & 68.66 & \textbf{74.19} & 71.32 & \textbf{93.13} & 91.18 & 92.14   \\
                  && PedCoT (ours)  & \textbf{85.33}  & \textbf{89.33}  & \textbf{90.00} & 86.36 & 61.29 & \textbf{71.70} & 90.62 & 97.48 & \textbf{93.93}   \\
        \hline
        \hline
        \multirow{12}{*}{\makecell[c]{PRM800K}} & \multirow{6}{*}{\makecell[c]{GPT-4}} & Direct-Step & 33.33 & 67.10 & 66.00 & 42.20 & 54.12 & 47.42 & 79.58 & 70.70 & 74.88 \\
                  && Vanilla & 37.00  &73.23  &73.00 & 52.22 & 55.29 & 53.71 & 81.90 & 80.00 & 80.94      \\
                  && Zero-shot CoT  & 41.33  & 74.89  & 74.00 & 53.15 & 69.41 & 60.20 & 86.24 & 75.81 & 80.69   \\
                  && Plan-and-Solve  & 38.00  & 74.79  & 74.00 & 53.27 & 67.06 & 59.37 & 85.49 & 76.74 & 80.88   \\
                  && SelfCheck  & 40.00  & 71.79  & 70.33 & 48.59 & \textbf{81.18} & 60.79 & \textbf{89.87} & 66.05 & 76.14    \\
                  && PedCoT (ours)  & \textbf{45.33}  & \textbf{79.82}  & \textbf{79.33} & \textbf{61.39} & 72.94 & \textbf{66.67} & 88.44 & \textbf{81.86} & \textbf{85.02}   \\
        \cline{2-12}
        & \multirow{6}{*}{\makecell[c]{GPT-4 Turbo}} & Direct-Step & 35.00 & 72.76 & 72.67 & 51.72 & 52.94 & 52.33 & 81.22 & 80.47 & 80.84 \\
                  && Vanilla & 34.00  &75.05  & 78.33 & 77.78 & 32.94 & 46.28 & 78.41 & \textbf{96.28} & 86.43      \\
                  && Zero-shot CoT  & 38.67  & 78.96  & 80.67 & 76.47 & 45.88 & 57.35 & 81.53 & 94.42 & 87.50   \\
                  && Plan-and-Solve  & 38.33  & 81.01  & 82.67 & \textbf{83.67} & 48.24 & 61.19 & 82.47 & \textbf{96.28} & \textbf{88.84}   \\
                  && SelfCheck  & 41.67  & 70.21  & 68.67 & 46.90 & \textbf{80.00} &  59.13 & 89.03 & 64.19 & 74.59  \\
                  && PedCoT (ours)  & \textbf{44.33}  & \textbf{83.45}  & \textbf{83.33} & 69.66 & 72.94 & \textbf{71.26} & \textbf{89.10} & 87.44 & 88.26   \\
        \bottomrule
    \end{tabular}
    \caption{Comparison of various CoT methods on the BIG-Bench Mistake and PRM800K datasets. $^\dagger$The scores are from the original paper.}
     \label{tab:main_result_bigbench}
\end{table*}

\paragraph{PRM800K}~\cite{prm800k}: The original dataset contains about 800,000 step-level labels over 75,000 solutions for math word problems in the MATH dataset~\cite{math_dataset}, which consists of 12,500 problems from high-school math competitions. The problems necessitate a comprehensive understanding of intricate mathematical concepts and problem-solving methodologies. We adopt this challenging dataset to verify the effectiveness and robustness of our method. We randomly select a set of 300 pairs of problems and their affiliated reasoning answer traces\footnote{Accessible on \url{github.com/HaoyuanPeng/PedCoT-IJCAI24/}}, containing 85 correct traces and 3,736 reasoning steps in total.
The steps in the PRM800K dataset are categorized as positive, negative, and neutral. To use the proposed method, we adjust the labels to align with our label schema. Specifically, the positive or neutral steps are treated as right (i.e., $L_{step}=1$) while the negative steps are labeled as wrong (i.e., $L_{step}=0$) in our experiments.


\subsection{Models, Baselines, and Metrics}

\paragraph{Models.} Our experiments are conducted with two state-of-the-art LLMs: GPT-4~\cite{gpt4} and its latest generation GPT-4 Turbo\footnote{We used gpt-4-1106-preview, the latest version then.}. The temperature for generation is consistently set to 0 for both models to minimize the diversity of model outputs. We do not employ any ensemble strategies, such as self-consistency~\cite{self-consistency}, to maintain a pure comparison with different CoT baseline methods.

\paragraph{Baselines.} We compare our proposed PedCoT with the following zero-shot prompting methods: (1) \textbf{Zero-shot CoT}~\cite{0shot-cot}; (2) \textbf{Plan-and-Solve Prompting}~\cite{plan-and-solve}; (3) \textbf{SelfCheck}~\cite{selfcheck}. We re-implement these baselines by prompting LLMs with the same prompts and multi-stage pipelines. To better evaluate the abilities of LLMs on detecting reasoning tasks with CoT prompting, we also consider \textbf{Direct-Step Prompting}~\cite{bigbenchmistake} and \textbf{Vanilla Two-stage Prompting} as our baselines. Direct-Step Prompting instructs LLMs to generate a single word \textit{yes} for each step if it is correct, or \textit{no} otherwise. Vanilla Two-stage Prompting first prompts LLMs to analyze each step's correctness without specific instructions. Subsequently, LLMs are prompted again to generate a label word based on their analyses. 



\paragraph{Metrics.} Similar to the previous work that evaluates the performance of mistake finding~\cite{bigbenchmistake}, we also utilize the mistake finding accuracy (abbr. MF. Acc.) and weighted average F1 (abbr. Avg. F1) scores as our metrics. 
Specifically, the MF. Acc. is strict for step-level accuracy which means mistakes should be found at the right step. Only the first mistake among a trace is concerned. The average F1 score provides a lenient measure of trace-level accuracy, indicating that mistakes within a trace scope can be identified. However, it does not account for whether these errors are pinpointed at the correct step. 
In addition, the binary classification accuracy (abbr. Cls. Acc.), as well as the precision ($P_+$ and $P_-$), recall ($R_+$ and $R_-$) and F1 ($F_+$ and $F_-$) for each label, are reported in the experiment. 

\subsection{Results and Analysis}


Table~\ref{tab:main_result_bigbench} reports the comparison among different zero-shot CoT methods applied with GPT-4 and GPT-4 Turbo on the two datasets. Consistent with previous research, directly prompting the LLMs to output a single word for the correctness of each reasoning step (i.e., Direct-Step) is ineffective even with the state-of-the-art LLMs. Vanilla Two-stage Prompting allows LLMs to generate several analyses before being prompted again for the final label word, leading to a slightly better performance than Direct-Step. Zero-shot CoT and Plan-and-Solve Prompting explicitly instruct LLMs to generate intermediate analyses, and show improvements compared to Vanilla Two-stage Prompting. The above results indicate that the more analysis contents before giving the final judgment are generated by prompting LLMs, the better performance is obtained, especially with GPT-4 Turbo.

Our PedCoT consistently outperforms all the other baselines by a remarkable margin on the two datasets and two LLMs. Particularly, on the BIG-Bench Mistake dataset, PedCoT surpasses SelfCheck across both Acc. and Avg. F1 metrics. On the PRM800K dataset, PedCoT achieves a higher MF. Acc. score ranging from 4.00\% to 7.33\% compared to two-stage CoT baselines when using GPT-4. Additionally, its Avg. F1 and Cls. Acc. scores surpass these baselines by about 4.93\% to 5.33\%. Compared to SelfCheck, PedCoT exhibits a 5.33\% leading in MF. Acc., and a considerable leading of 8.03\% and 9.00\% in Avg. F1 and Cls. Acc. scores respectively. When using GPT-4 Turbo, PedCoT exhibits a MF. Acc. advantage of 5.66\% compared to the two-stage CoT baselines, along with improvements in Avg. F1 and Cls. Acc. In comparison to SelfCheck, PedCoT achieves a higher MF. Acc. by 2.66\%, and a leading of 13.24\% and 14.66\% in Avg. F1 and Cls. Acc. respectively. All of the above results demonstrate the effectiveness of leveraging pedagogical knowledge in our proposed methods. 



Additionally, although SelfCheck outperforms other baselines on BIG-Bench Mistake dataset, as displayed in Table~\ref{tab:main_result_bigbench}, its Avg. F1 and Cls. Acc. scores on PRM800K dataset fall behind, as shown in Table~\ref{tab:main_result_bigbench}. The results may be caused by the following reasons: During the process of SelfCheck prompting, one of the LLM requests is to summarize the target of the current answer step, and then regenerate a new current step for subsequent comparison. For the BIG-Bench Mistake dataset, due to the easy arithmetic ability required to solve math problems, SelfCheck can perform well. However, the PRM800K dataset presents more complex math problems requiring advanced reasoning skills. In such cases, SelfCheck may summarize an incorrect target, leading to a incidental alignment with an incorrect actual current step. This bias in SelfCheck can result in erroneously classifying the real current step as correct, impacting the precision for correct traces ($P_+$) and reducing recall for incorrect traces ($R_-$). The reason of our method can surmount these challenges met with SelfCheck, by our analysis, is that the PedCoT is grounded in pedagogical principles and allows LLMs to remain blind to the current step. This approach enables the LLMs to render judgments from an impartial and unbiased perspective.



\subsection{Ablation Study}

We design ablation experiments to assess the contribution of each principle. Specifically, we deactivate the three principles one by one, and observe the performance change patterns of our proposed methods. In this study, deactivating a principle means that this principle's corresponding prompt contents are removed. We select 25\% of the test set for ablation experiments. The results are reported in Table~\ref{tab:ablation}.

\paragraph{Ablating prompt contents of Principle 1 (Remember).} We observe a slight decrease in metrics when Principle 1 is deactivated, which is about mathematical concepts. Hence, a positive contribution of Principle 1 on both BIG-Bench Mistake dataset and PRM800K dataset is validated. Meanwhile, the experimental results on the BIG-Bench Mistake dataset exhibit a relatively smaller decrease, which leads us to believe the easier arithmetic problems of BIG-Bench Mistake dataset require only the concept of the rules of arithmetic.

\paragraph{Ablating prompt contents of Principle 2 (Understand).} We observe some fluctuations in the performance of the proposed method on BIG-Bench Mistake, while there is a slight decrease in performance on the PRM800K dataset after deactivating Principle 2, which is about problem-solving approaches. The observed variations can also be attributed to the varying complexities of the problems. Our analysis suggests that the strategies required to solve arithmetic problems are less intricate compared to those needed for complex mathematical word problems. 


\begin{table}
    \centering
    \resizebox{\linewidth}{!}{
    \begin{tabular}{llccc|ccc}
        \toprule
        \multirow{2}{*}{\makecell[c]{LLMs}} & \multirow{2}{*}{\makecell[c]{Methods}} & \multicolumn{3}{c|}{BIG-Bench Mistake} & \multicolumn{3}{c}{PRM800K} \\ 
               & & MF. Acc. & Avg. F1 & Cls. Acc. & MF. Acc. & Avg. F1 & Cls. Acc. \\
        \hline
        \multirow{4}{*}{\makecell[c]{GPT-4}} & PedCoT  & 90.67 & 91.78 & 92.00 & 46.67 & 77.87 & 77.33 \\
          & -Principle 1 & 89.33 & 90.54 & 90.67 & 42.67 & 74.06 & 73.33 \\
          & -Principle 2 & 89.33 & 90.54 & 90.67 & 42.67 & 76.00 & 76.00 \\
          & -Principle 3 & 28.00 & 37.44 & 36.00 & 32.00 & 57.92 & 57.33 \\

        \hline
        \multirow{4}{*}{\makecell[c]{GPT-4\\Turbo}} & PedCoT & 92.00 & 91.51 & 92.00 & 53.33 & 84.32 & 84.00 \\
        & -Principle 1 & 90.67 & 91.18 & 92.00 & 52.00 & 82.77 & 82.67 \\
        & -Principle 2 & 90.67 & 92.00 & 92.00 & 52.00 & 84.18 & 82.67 \\
        & -Principle 3 & 21.33 & 32.22 & 30.67 & 36.00 & 62.29 & 61.33 \\
        \bottomrule
    \end{tabular}
    }
    \caption{Ablation experiment of PedCoT by ablating one principle.}
     \label{tab:ablation}
\end{table}
\paragraph{Ablating prompt contents of Principle 3 (Apply).} When the portion of the prompt representing Principle 3, which is about calculations, is removed from the prompt, there is a significant decline in the performance of the proposed method across both datasets and both LLMs. This indicates that verifying the correctness of calculations, including mathematical expression transformations and simplifications, is the most fundamental and crucial ability required for prompting the LLMs to detect mathematical reasoning errors.

Since the datasets have varying complexities of mathematical problems, the performance change patterns are not always significant.
Regardless the performance stays stable or shows a significant decrease, it can be concluded that each principle has a positive impact on the PedCoT method. This demonstrates the effectiveness of incorporating the pedagogical theory of each \textit{learning ability} from BCM.

\subsection{Study on One-Stage vs. Two-Stage Prompting}

In this section, we investigate the necessity of two-stage prompting. A distinctive feature of our two-stage methodology is that, at Stage-1, the current step remains undisclosed to LLMs and thus they can make an unbiased judgement at Stage-2. To verify the necessity, we construct a version of one-stage prompt by combining our two-stage prompts.



We compare the performance between the two-stage PedCoT and its one-stage variant, utilizing two datasets with GPT-4. As presented in Table~\ref{tab:comparison_1stage}, the precision of correct traces ($P_+$) and the recall of incorrect traces ($R_-$) significantly degrade when employing one-stage prompting. This suggests that the LLMs would be misled by the real current step present in the one-stage prompt. Based on our analysis, it appears that when the real current step is shared with the LLMs, they may encounter difficulties in independently generating analyses. Instead, they seem to rely heavily on referencing the real current step. We call this phenomenon `lazy LLMs' since LLMs are inclined to cater to the ideas present in prompts. Consequently, finding mistakes by giving the real current step becomes difficult to accomplish. This experiment demonstrates that our two-stage prompting is necessary.

\begin{table}
    \centering
    \resizebox{\linewidth}{!}{
    \begin{tabular}{llcccc}
        \toprule
        Dataset  & Strategy & MF. Acc. & Cls. Acc. & $P_{+}$ & $R_{-}$ \\
        \midrule
        \multirow{2}{*}{\makecell[c]{BIG-Bench Mistake}} & PedCoT & 83.00   & 89.33 & 81.25 & 96.22     \\
                  & One-Stage & 74.33 & 81.00  & 53.33 & 85.29  \\
        \hline
        \multirow{2}{*}{\makecell[c]{PRM800K}} & PedCoT  & 45.33   & 79.33 & 61.39 & 81.86    \\
                  & One-Stage & 35.00 & 61.00 & 38.41 & 60.47  \\
    \bottomrule
    \end{tabular}
    }
    \caption{Two-stage PedCoT vs. its one-stage variant on GPT-4.}
     \label{tab:comparison_1stage}
\end{table}

\section{Conclusion}

In this paper, we investigate a novel problem of how to leverage domain knowledge of pedagogy to enhance the complex reasoning abilities of the zero-shot prompted LLMs.
We learn from pedagogical theories to instruct LLMs and to guide the prompts design for LLMs. We propose a novel strategy, namely PedCoT method, which consists of three pedagogy principles for prompt design, two-stage interaction process, and pedagogical Chain-of-Thought prompts. We experiment two public datasets of math problems with various complexity levels. The state-of-the-art performance of our proposed method indicates that it can achieve efficient reasoning mistaking finding with only two requests to LLMs. In addition, the ablation experiment and study on one-stage method validate the importance of the three pedagogical principles and the necessity of two-stage prompting. Overall, this finding indicates that LLMs actually can find mathematical reasoning mistakes by resorting to domain knowledge.





\section*{Acknowledgments}

This work was supported by the Humanities and Social Science Fund of Ministry of Education (No. 23YJC790187). We thank the anonymous reviewers for their insightful and constructive comments.








\bibliographystyle{named}
\bibliography{ijcai24}

\onecolumn
\section{Appendix}
\appendix 

\section{Using PedCoT on GPT-3.5-Turbo}
 
Table~\ref{tab:main-result-gpt-3.5-turbo} shows the comparison between PedCoT prompting and Direct-Step prompting with GPT-3.5-Turbo. PedCoT outperforms Direct-Step prompting by a significant margin on both datasets. Particularly, on the BIG-Bench Mistake dataset, PedCoT improves the mistake finding accuracy from 26.00\% to 75.67\%, and the average F1 from 64.08\% to 87.20\%. On the PRM800K dataset, PedCoT achieved improvements of 16.67\%, 13.20\%, and 13.33\% in MF. Acc., Avg. F1, and Cls. Acc. respectively. This experiment can indicate that our PedCoT method is also effective with an earlier low-version LLM.

\begin{table*}[h]
    \centering
    \resizebox{\linewidth}{!}{
    \begin{tabular}{llccccccccc}
        \toprule
        Datasets  & CoT methods & MF. Acc. & Avg. F1 & Cls. Acc. & $P_{+}$ & $R_{+}$ & $F _{+}$ & $P_{-}$ & $R_{-}$ & $F_{-}$ \\
        \midrule
        \multirow{2}{*}{\makecell[c]{BIG-Bench Mistake}} & Direct-Step & 26.00 & 64.08 & - & - & - & - & - & - & - \\
                  & PedCoT (ours)  & \textbf{75.67}  & \textbf{87.20}  & 88.00 & 79.55 & 56.45 & 66.04 & 89.45 & 96.22 & 92.71  \\
        \hline
        \multirow{2}{*}{\makecell[c]{PRM800K}} & Direct-Step & 20.00 & 63.04 & 62.67 & 35.16 & 37.65 & 36.36 & 74.64 & 72.56 & 73.58 \\
                  
                  & PedCoT (ours)  & \textbf{36.67}  & \textbf{76.24}  & \textbf{76.00} & \textbf{57.14} & \textbf{61.18} & \textbf{59.09} & \textbf{84.21} & \textbf{81.86} & \textbf{83.02}   \\
        \bottomrule
    \end{tabular}
    }
    \caption{Comparison of our PedCoT with baseline when using GPT-3.5-Turbo.}
    \label{tab:main-result-gpt-3.5-turbo}
\end{table*}

\section{The Complete PedCoT Prompts and LLM Responses of the Example in Main Body's Figure 2}

We present a complete two-stage interaction process (TIP) of PedCoT strategy with a LLM (e.g., GPT-4 Turbo) in Table~\ref{appendix:example}. 
The background colors are consistent with main body's Figure 2, that is, the texts with a \colorbox[RGB]{202,199,199}{grey} background represent the instruction template $T_{j}$, the texts with a \colorbox[RGB]{235, 205, 252}{purple} background represent the question $Q$, the texts with a \colorbox[RGB]{178, 253, 177}{green} background represent the answer steps $\{a_i\}$, the texts with a \colorbox[RGB]{255,226,151}{yellow} background represent the outputs $\{G^{(k)}\}$ from the LLM at Stage-1, the texts with a \colorbox[RGB]{161,252,247}{blue} background represent the outputs $\{E^{(k)}\}$ extracted by the LLM at Stage-2, and the texts with a \colorbox[RGB]{255,164,174}{red} background represent the comparison and principle mistake labels $L_{principle}^{(k)}$, where $j$ means the $j$-th stage, $i$ means the $i$-th answer step, and $k$ corresponds to the $k$-th pedagogical principle. The uncolored texts are just notes to segment the input and output of two stages and are not part of either the input prompts or LLM responses.

We sequentially split the Stage-1 response $G$ to $k$ segments corresponding to each pedagogical principle with \textit{\underline{\textbackslash n\textbackslash n3.}}, \textit{\underline{\textbackslash n\textbackslash n2.}} and \textit{\underline{\textbackslash n\textbackslash n1.}} to obtain $G^{(1)}$, $G^{(2)}$ and $G^{(3)}$, and then integrate them into the instruction template $T_2$ to construct the input prompt $P_2$ for Stage-2.

\begin{longtable}{p{\textwidth}}
    \label{appendix:example} \\
    \toprule
    \toprule

\textit{Stage-1 Prompt: $P_1=\{Q,A_{i-1}, T_1\}$} \\

\colorbox[RGB]{202,199,199}{\parbox{\textwidth}{You are given a math problem and several initial steps of the solution process.}} \\

\colorbox[RGB]{235, 205, 252}{\parbox{\textwidth}{Question: Let $f(x) = (\arccos x)^2 + (\arcsin x)^2.$ Find the range of $f(x).$  All functions are in radians.}} \\

\colorbox[RGB]{202,199,199}{\parbox{\textwidth}{Initial steps:}} \\
\colorbox[RGB]{178, 253, 177}{\parbox{\textwidth}{(step 1) I notice that the domain of both $\arccos$ and $\arcsin$ is the interval $[-1, 1]$, so I can start by assuming that $x$ is in that interval.}} \\

\colorbox[RGB]{202,199,199}{\parbox{\textwidth}{Execute the following instructions sequentially. \\

1. List the mathematical concepts that should be applied in the next step for solving the problem. \\

2. List the key analyses that should be derived in the next step for solving the problem. \\

3. List the mathematical expressions that have not been fully computed yet in the question and the initial steps. Then carefully calculate their corresponding results step by step.}} \\ 

\midrule
\textit{Stage-1 Response: $G=\{G^{(1)},G^{(2)},G^{(3)}\}$} \\

\colorbox[RGB]{255,226,151}{\parbox{\textwidth}{1. Mathematical Concepts to Apply:}} \\
   \colorbox[RGB]{255,226,151}{\parbox{\textwidth}{- Trigonometric identities, specifically the Pythagorean identity: $\sin^2(\theta) + \cos^2(\theta) = 1$.}} \\
   \colorbox[RGB]{255,226,151}{\parbox{\textwidth}{- The fact that $\arcsin(x)$ and $\arccos(x)$ are inverse functions of $\sin(\theta)$ and $\cos(\theta)$, respectively, and that for a given value of $x$, $\sin(\arcsin(x)) = x$ and $\cos(\arccos(x)) = x$.}} \\
   \colorbox[RGB]{255,226,151}{\parbox{\textwidth}{- The properties of the $\arcsin$ and $\arccos$ functions, including their ranges: $\arcsin(x) \in [-\frac{\pi}{2}, \frac{\pi}{2}]$ and $\arccos(x) \in [0, \pi]$.}} \\
   \colorbox[RGB]{255,226,151}{\parbox{\textwidth}{- Calculus concepts may be used to find the extrema of the function $f(x)$ if necessary.}} \\

\colorbox[RGB]{255,226,151}{\makebox[\textwidth][l]{\strut\hspace{0pt}}} \\

\colorbox[RGB]{255,226,151}{\parbox{\textwidth}{2. Key Analyses for the Next Step:}} \\
\colorbox[RGB]{255,226,151}{\parbox{\textwidth}{- Recognize the relationship between the functions $\arccos(x)$ and $\arcsin(x)$.}} \\
\colorbox[RGB]{255,226,151}{\parbox{\textwidth}{- Use the Pythagorean identity for sine and cosine to relate $\arccos(x)$ and $\arcsin(x)$.}} \\
\colorbox[RGB]{255,226,151}{\parbox{\textwidth}{- Determine the maximum and minimum values of the function $f(x)$ within the domain of $x \in [-1, 1]$.}} \\

\colorbox[RGB]{255,226,151}{\makebox[\textwidth][l]{\strut\hspace{0pt}}} \\

\colorbox[RGB]{255,226,151}{\parbox{\textwidth}{3. Mathematical Expressions to Compute:}} \\
\colorbox[RGB]{255,226,151}{\parbox{\textwidth}{- We need to compute the expression for $f(x) = (\arccos x)^2 + (\arcsin x)^2$ using the relationship between $\arccos(x)$ and $\arcsin(x)$.}} \\

\colorbox[RGB]{255,226,151}{\parbox{\textwidth}{Let's compute the expression step by step:}} \\

\colorbox[RGB]{255,226,151}{\parbox{\textwidth}{Since $\arccos(x)$ and $\arcsin(x)$ are related by the identity $\arccos(x) + \arcsin(x) = \frac{\pi}{2}$ for $x \in [-1, 1]$, we can use this to simplify $f(x)$.}} \\

\colorbox[RGB]{255,226,151}{\parbox{\textwidth}{We have:
\[f(x) = (\arccos x)^2 + (\arcsin x)^2\]
\[f(x) = \left(\frac{\pi}{2} - \arcsin(x)\right)^2 + (\arcsin x)^2\]
\[f(x) = \frac{\pi^2}{4} - \pi \arcsin(x) + (\arcsin(x))^2 + (\arcsin x)^2\]
\[f(x) = \frac{\pi^2}{4} - \pi \arcsin(x) + 2(\arcsin(x))^2\] }}\\ 

\midrule
\midrule
\textit{Stage-2 Prompt: $P_2=\{Q,A_i,G,T_2\}$} \\

\colorbox[RGB]{202,199,199}{\parbox{\textwidth}{You are given a math problem and several initial steps of the solution process.}} \\

\colorbox[RGB]{235, 205, 252}{\parbox{\textwidth}{Question: Let $f(x) = (\arccos x)^2 + (\arcsin x)^2.$ Find the range of $f(x).$  All functions are in radians.}} \\

\colorbox[RGB]{202,199,199}{\parbox{\textwidth}{Initial steps:}} \\
\colorbox[RGB]{178, 253, 177}{\parbox{\textwidth}{(step 1) I notice that the domain of both arccos and arcsin is the interval [-1, 1], so I can start by assuming that x is in that interval.}} \\

\colorbox[RGB]{202,199,199}{\parbox{\textwidth}{The actual next step is:}} \\
\colorbox[RGB]{178, 253, 177}{\parbox{\textwidth}{(step 2) Next, I recall that arccos and arcsin are inverse functions, meaning that arccos(arcsin(x)) = x and arcsin(arccos(x)) = x for any x in [-1, 1].}} \\

\colorbox[RGB]{202,199,199}{\parbox{\textwidth}{Execute the following instructions sequentially. \\
1. First extract the mathematical concepts employed in the actual next step. Below is the correct mathematical concepts should be applied in the next step. Give a detailed comparison between the extracted concepts and the correct mathematical concepts.}} \\

\colorbox[RGB]{255,226,151}{\parbox{\textwidth}{Mathematical Concepts to Apply:}} \\
   \colorbox[RGB]{255,226,151}{\parbox{\textwidth}{- Trigonometric identities, specifically the Pythagorean identity: $\sin^2(\theta) + \cos^2(\theta) = 1$.}} \\
   \colorbox[RGB]{255,226,151}{\parbox{\textwidth}{- The fact that $\arcsin(x)$ and $\arccos(x)$ are inverse functions of $\sin(\theta)$ and $\cos(\theta)$, respectively, and that for a given value of $x$, $\sin(\arcsin(x)) = x$ and $\cos(\arccos(x)) = x$.}} \\
   \colorbox[RGB]{255,226,151}{\parbox{\textwidth}{- The properties of the $\arcsin$ and $\arccos$ functions, including their ranges: $\arcsin(x) \in [-\frac{\pi}{2}, \frac{\pi}{2}]$ and $\arccos(x) \in [0, \pi]$.}} \\
   \colorbox[RGB]{255,226,151}{\parbox{\textwidth}{- Calculus concepts may be used to find the extrema of the function $f(x)$ if necessary.}} \\

\colorbox[RGB]{202,199,199}{\parbox{\textwidth}{Finally, output a label from the options (correct-and-aligned, reasonable-but-incomplete, nothing-extracted, and contradiction-found) to categorize the use of mathematical concepts for the actual next step.}} \\

\colorbox[RGB]{202,199,199}{\parbox{\textwidth}{2. First extract the key analyses present in the actual next step. Below is the correct analyses should be derived in the next step. Give a detailed comparison between the actual analyses and the correct analyses.}} \\

\colorbox[RGB]{255,226,151}{\parbox{\textwidth}{Key Analyses for the Next Step:}} \\
\colorbox[RGB]{255,226,151}{\parbox{\textwidth}{- Recognize the relationship between the functions $\arccos(x)$ and $\arcsin(x)$.}} \\
\colorbox[RGB]{255,226,151}{\parbox{\textwidth}{- Use the Pythagorean identity for sine and cosine to relate $\arccos(x)$ and $\arcsin(x)$.}} \\
\colorbox[RGB]{255,226,151}{\parbox{\textwidth}{- Determine the maximum and minimum values of the function $f(x)$ within the domain of $x \in [-1, 1]$.}} \\

\colorbox[RGB]{202,199,199}{\parbox{\textwidth}{Finally, output a label from the options (correct-and-aligned, reasonable-but-incomplete, nothing-extracted, and contradiction-found) to categorize the problem solving approach for the actual next step.}} \\

\colorbox[RGB]{202,199,199}{\parbox{\textwidth}{3. First extract the calculation results present in the actual next step.
Below is the correct calculation to be done in the next step. If any calculation results are extracted from the actual next step, list their corresponding results in the correct calculations, and give a detailed comparison between the extracted calculation results and correct results.}}

\colorbox[RGB]{255,226,151}{\parbox{\textwidth}{Mathematical Expressions to Compute:}} \\
\colorbox[RGB]{255,226,151}{\parbox{\textwidth}{- We need to compute the expression for $f(x) = (\arccos x)^2 + (\arcsin x)^2$ using the relationship between $\arccos(x)$ and $\arcsin(x)$.}} \\

\colorbox[RGB]{255,226,151}{\parbox{\textwidth}{Let's compute the expression step by step:}} \\

\colorbox[RGB]{255,226,151}{\parbox{\textwidth}{Since $\arccos(x)$ and $\arcsin(x)$ are related by the identity $\arccos(x) + \arcsin(x) = \frac{\pi}{2}$ for $x \in [-1, 1]$, we can use this to simplify $f(x)$.}} \\

\colorbox[RGB]{255,226,151}{\parbox{\textwidth}{We have:
\[f(x) = (\arccos x)^2 + (\arcsin x)^2\]
\[f(x) = \left(\frac{\pi}{2} - \arcsin(x)\right)^2 + (\arcsin x)^2\]
\[f(x) = \frac{\pi^2}{4} - \pi \arcsin(x) + (\arcsin(x))^2 + (\arcsin x)^2\]
\[f(x) = \frac{\pi^2}{4} - \pi \arcsin(x) + 2(\arcsin(x))^2\] }}\\ 

\colorbox[RGB]{202,199,199}{\parbox{\textwidth}{Finally, output a label from the options (correct-and-aligned, reasonable-but-incomplete, nothing-extracted, and contradiction-found) to categorize the correctness of calculations for the actual next step.}} \\

\midrule

\textit{Stage-2 Response: $E=\{E^{(1)},E^{(2)},E^{(3)}\}$ and $L_{principle}^{(k)}$} \\

\colorbox[RGB]{161,252,247}{\parbox{\textwidth}{1. Mathematical Concepts to Apply:}} \\
\colorbox[RGB]{161,252,247}{\parbox{\textwidth}{The actual next step does not employ the correct mathematical concepts. The correct concepts involve using the Pythagorean identity and understanding the properties of the arcsin and arccos functions, including their ranges. The actual next step does not mention the Pythagorean identity or the ranges of the functions.}} 

\colorbox[RGB]{255,164,174}{\parbox{\textwidth}{Therefore, the label for the use of mathematical concepts is "contradiction-found."}} \\

\colorbox[RGB]{161,252,247}{\parbox{\textwidth}{2. Key Analyses for the Next Step:}} \\ 
\colorbox[RGB]{161,252,247}{\parbox{\textwidth}{The actual next step incorrectly states that arccos and arcsin are inverse functions in a way that suggests $arccos(arcsin(x)) = x$ and $arcsin(arccos(x)) = x$ for any $x$ in $[-1, 1]$, which is not true. The correct relationship is that for a given angle $\theta$, $sin(arccos(\theta))$ and $cos(arcsin(\theta))$ will give the corresponding leg lengths of a right triangle where $\theta$ is one of the non-right angles. The correct analyses should recognize the relationship between $arccos(x)$ and $arcsin(x)$ and use the Pythagorean identity to relate them.}}
\colorbox[RGB]{255,164,174}{\parbox{\textwidth}{ The actual next step does not do this, and therefore, the label for the problem-solving approach is "contradiction-found."}}\\

\colorbox[RGB]{161,252,247}{\parbox{\textwidth}{3. Mathematical Expressions to Compute:}} \\
\colorbox[RGB]{161,252,247}{\parbox{\textwidth}{No calculations are extracted from the actual next step. The correct calculation involves using the identity $arccos(x) + arcsin(x) = \pi/2$ and then finding the range of $f(x)$ by considering the maximum and minimum values of $arcsin(x)$ within its domain.}}
\colorbox[RGB]{255,164,174}{\parbox{\textwidth}{Since the actual next step does not provide any calculation results, the label for the correctness of calculations is "nothing-extracted."}} \\

\bottomrule
\bottomrule
\caption{An example of the complete prompt texts and LLM responses, corresponding to the main body's Figure 2.}
\end{longtable}

\end{document}